\documentclass[11pt]{article}
\usepackage{coling2016}
\usepackage{times}
\usepackage{url}
\usepackage{latexsym}
\usepackage{amsmath}
\usepackage{graphicx}
\usepackage[table,xcdraw]{xcolor}
\usepackage{wrapfig}

\graphicspath{ {images/} }
\usepackage{amssymb}
\usepackage{CJKutf8}
\usepackage{multirow}
\usepackage{dirtytalk}
\usepackage{booktabs}
\usepackage{xspace}  
\usepackage{array}
\usepackage{diagbox}
\newcolumntype{P}[1]{>{\centering\arraybackslash}p{#1}}

\newcommand{\etal}{et~al\mbox{.}\xspace}  %

\title{UTCNN: a Deep Learning Model of Stance Classification 
\\on Social Media Text}

\author{Wei-Fan Chen \\
  Institute of Information Science, \\
  Academia Sinica, Taipei, Taiwan. \\
  {\tt viericwf@iis.sinica.edu.tw} \\\And
  Lun-Wei Ku \\
  Institute of Information Science, \\
  Academia Sinica, Taipei, Taiwan.\\
  {\tt lwku@iis.sinica.edu.tw} \\}

\date{}

\begin{document}
\maketitle
\begin{abstract}
Most neural network models for document classification on social media
focus on text information to the neglect of other information on these platforms.
In this paper, we classify post stance on
social media channels and develop UTCNN, a neural network model that incorporates
user tastes, topic tastes, and user comments on posts. UTCNN not only
works on social media texts, but also analyzes texts in
forums and message boards. Experiments performed on Chinese Facebook
data and English online debate forum data show that UTCNN
achieves a 0.755 macro-average f-score for supportive, neutral, and
unsupportive stance classes on Facebook data, which is significantly better than
models in which either user, topic, or comment information is withheld. This
model design greatly mitigates the lack of data for the minor class without the use of oversampling. 
In addition, UTCNN yields a 0.842 accuracy on English online debate forum data, which
also significantly outperforms results from previous work as well as other deep learning models, showing that UTCNN
performs well regardless of language or platform.
\end{abstract}

\section{Introduction}
\blfootnote{
    
    \hspace{-0.65cm}  
    This work is licenced under a Creative Commons 
    Attribution 4.0 International License.
    License details:
    \url{http://creativecommons.org/licenses/by/4.0/}
}
Deep neural networks have been widely used in text classification and have achieved
promising results \cite{lai2015recurrent,ren2016context,huang2016modeling}.
Most focus on content information and use models
such as convolutional neural networks (CNN) \cite{CNN} or recursive
neural networks \cite{socher2013recursive}. However, for user-generated
posts on social media like Facebook or Twitter, there is more information that
should not be ignored.
On social media platforms, a user can act either as the author of a post or as a
reader who expresses his or her comments about the post. 

In this paper, we classify posts taking into account post authorship, 
likes, topics, and comments.
In particular, users and their ``likes'' hold strong
potential for text mining. For example, given a set of posts that are related
to a specific topic, a user's likes and dislikes provide clues for stance
labeling. From
a user point of view, users with positive attitudes toward the issue
leave positive comments on the posts with praise or even just the post's content;
from a post point of view, positive posts attract users who hold positive
stances. We also investigate the influence of topics: 
different topics are associated with different stance labeling tendencies and word usage. 
For example we discuss women's rights and unwanted babies on the topic of 
abortion, but we criticize medicine usage or crime when on the topic of marijuana
\cite{hasan2014you}. Even for posts on a specific topic like
nuclear power, a variety of arguments are raised: green energy, radiation,
air pollution, and so on. As for comments, we treat them as additional text
information. The arguments in the comments and the commenters (the users who
leave the comments) provide hints on the post's content and further facilitate
stance classification.

In this paper, we propose the user-topic-comment neural network (UTCNN), a deep 
learning model that utilizes user, topic, and comment information. We attempt to
learn user and topic representations which encode user interactions and
topic influences to further enhance text classification, and we also incorporate
comment information. We evaluate this model on a post stance classification
task on forum-style social media platforms. The contributions of this paper are
as follows:
1. We propose UTCNN, a neural network for text in modern
social media channels as well as legacy social media, forums, and message
boards --- anywhere that reveals users, their tastes, as well as their replies to posts.  
2. When classifying social media post stances, we leverage users, including authors and likers. 
User embeddings can be generated even for users who have never posted anything.
3. We incorporate a topic model to automatically assign topics to each post in 
a single topic dataset.
4. We show that overall, the proposed method achieves the highest
performance in all instances, and that all of the information extracted, 
whether users, topics, or comments, still has its contributions. 

\section{Related Work} 
\subsection{Extra-Linguistic Features for Stance Classification}
In this paper we aim to use text as well as other features to see how they
complement 
each other in a deep learning model. In the stance classification domain,
previous work has showed that text features are limited, suggesting 
that adding extra-linguistic constraints could improve performance
\cite{bansal2008power,hasan2013extra,walker2012stance}. For example, Hasan and
Ng as well as Thomas \etal require that posts written by the same
author have the same stance \cite{hasan2013stance,thomas2006get}. 
The addition of this constraint yields accuracy improvements of 1--7\% for some models 
and datasets.  
Hasan and Ng later added user-interaction constraints and ideology constraints
\cite{hasan2013extra}: the former models the relationship among posts in a
sequence of replies and the latter models inter-topic relationships, e.g.,
users who oppose abortion could be conservative and thus are likely to oppose gay rights.

For work focusing on online forum text, since posts are linked through
user replies, sequential labeling methods have been used to model
relationships between posts. For example, Hasan and Ng use hidden Markov
models (HMMs) to model dependent relationships to the preceding post
\cite{hasan2013stance}; Burfoot \etal use iterative classification to
repeatedly generate new estimates based on 
the current state of knowledge 
\cite{burfoot2011collective}; Sridhar \etal use probabilistic soft logic (PSL)
to model reply links via collaborative filtering \cite{sridhar2015joint}. In
the Facebook dataset we study, we use comments instead of reply links.
However, as the ultimate goal in this paper is predicting not comment stance but 
post stance, we treat comments as extra information for use in predicting post stance.

\subsection{Deep Learning on Extra-Linguistic Features}
In recent years neural network models have been applied to document sentiment classification
\cite{socher2012semantic,socher2013recursive,kalchbrenner2014convolutional,johnson2014effective,huang2016modeling}. 
Text features can be used in deep networks to capture text semantics or sentiment.
For example, Dong \etal use an adaptive
layer in a recursive neural network for target-dependent Twitter sentiment
analysis, where targets are topics such as \emph{windows 7} or \emph{taylor
swift} \cite{adaptive,dong2014adaptive}; recursive neural tensor networks (RNTNs)
utilize sentence parse trees to capture sentence-level sentiment 
for movie reviews \cite{socher2013recursive}; Le and Mikolov predict sentiment by using
paragraph vectors to model each paragraph as a continuous representation
\cite{le2014distributed}. They show that performance can thus be improved by
more delicate text models.

Others have suggested using extra-linguistic features to improve
the deep learning model. The user-word composition vector model (UWCVM)
\cite{tang2015user} is inspired by the possibility that the strength of sentiment
words is user-specific; to capture this they add user embeddings in their
model. In UPNN, a later extension, they further add a product-word
composition as product embeddings, arguing that products can also show 
different tendencies of being rated or reviewed \cite{UPNN}. Their addition of
user information yielded 2--10\% improvements in accuracy as compared to the 
above-mentioned RNTN and paragraph vector methods. We also seek to inject user
information into the neural network model. In comparison to the research of Tang \etal on
sentiment classification for product reviews, the difference is two-fold.
First, we take into account multiple users (one author and potentially
many likers) for one post, whereas only one user (the reviewer) is involved in a
review. Second, we add comment information to provide more
features for post stance classification. None of these two factors have been
considered previously in a deep learning model for text stance classification.
Therefore, we propose UTCNN, which generates and utilizes user embeddings for
all users~---~even for those who have not authored any posts~---~and incorporates
comments to further improve performance.

\section{Method}
In this section, we first describe CNN-based document composition, which
captures user- and topic-dependent document-level semantic
representation from word representations. Then we show how to add comment
information to construct the user-topic-comment neural network (UTCNN).

\subsection{User- and Topic-dependent Document Composition}
\begin{figure}[tb]
\centering
\includegraphics[clip, trim=0.5cm 4.8cm 1.5cm 9.4cm, width=0.6\textwidth]{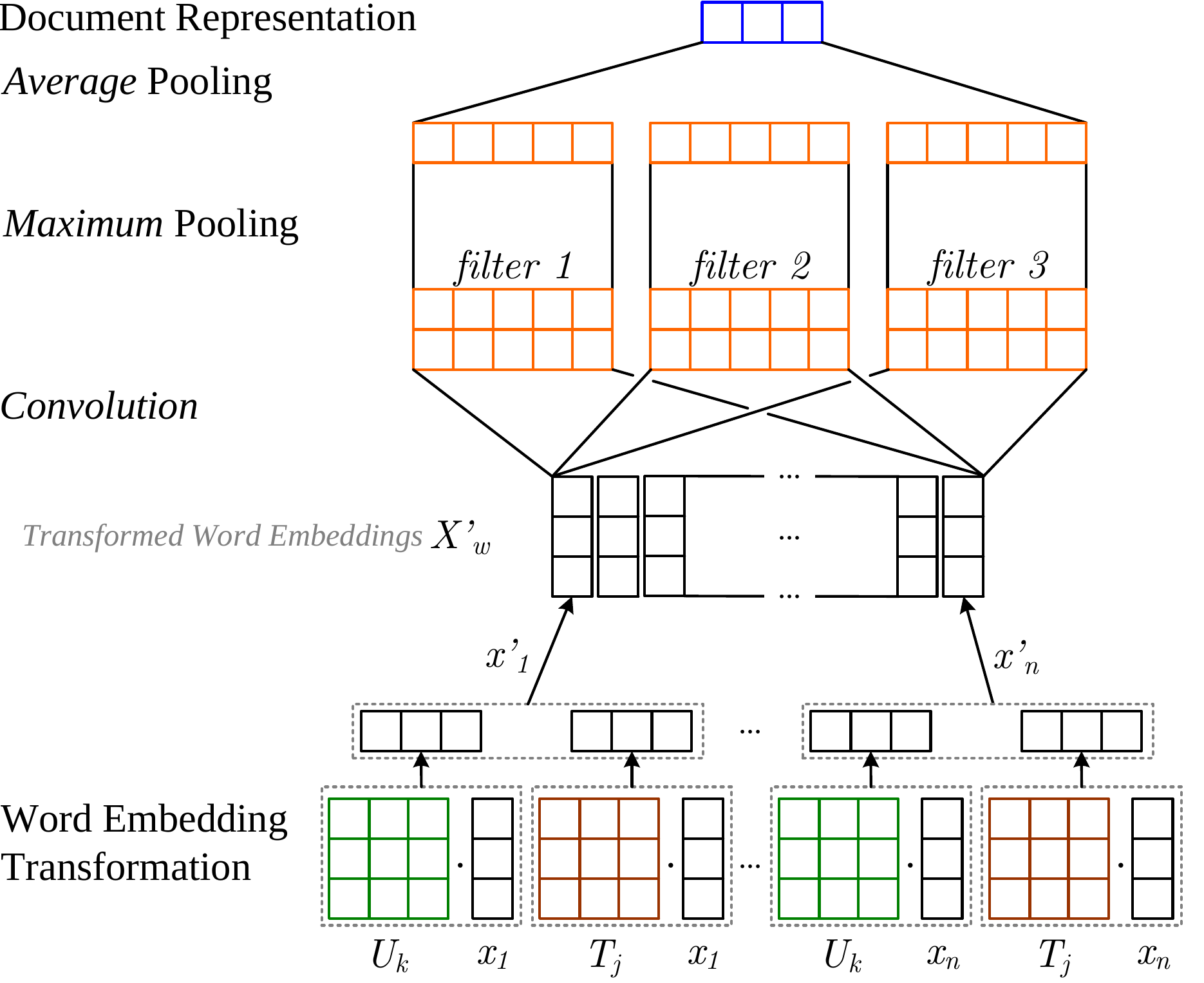}
\caption{Document composition in a convolutional neural network with three
convolutional filters and user- and topic-dependent semantic
transformations. Respectively, $x_w$ is the word embedding of word $w$,
$x'_w$ is the word embedding of word $w$ after transformation,
$U_k$ and $T_j$ are user and topic matrix embeddings for user $k$ and topic $j$.}
\label{fig:dc}
\end{figure}
As shown in Figure~\ref{fig:dc}, we use a general CNN \cite{CNN} and two
semantic transformations for document composition \footnote{Here by saying \textit{document}, we mean the user-generated content in a post or a comment.} . 
We are given a document with an engaged user $k$, a topic $j$, and its composite
$n$ words, each word $w$ of which is associated with a word embedding $x_w \in
\mathbb{R}^d$ where $d$ is the vector dimension. For each word embedding $x_w$,
we apply two dot operations as shown in Equation~\ref{eq:dot}:
\begin{equation}
{x'_w} = [U_k \cdot {x_w}; T_j \cdot {x_w} ]
\label{eq:dot}
\end{equation}
where $U_k \in \mathbb{R}^{d_u \times d}$  models the user reading preference
for certain semantics, and $T_j \in \mathbb{R}^{d_t \times d}$ models the
topic semantics; $d_u$ and $d_t$ are the dimensions of transformed user and
topic embeddings respectively. We use $U_k$ to model semantically what each user
prefers to read and/or write, and use $T_j$ to model the semantics of each topic.
The dot operation of $U_k$ and $x_w$ transforms the global representation $x_w$
to a user-dependent representation. Likewise, the dot operation of $T_j$ and $x_w$
transforms $x_w$ to a topic-dependent representation. 

After the two dot operations on $x_w$, we have user-dependent and
topic-dependent word vectors $U_k \cdot {x_w}$ and $T_j \cdot {x_w}$, which are
concatenated to form a user- and topic-dependent word vector $x'_w$. Then
the transformed word embeddings $X'_w=[x'_1;x'_2;...;x'_n]$ are used as the
CNN input.
Here we apply three convolutional
layers on the concatenated transformed word embeddings $x'_c = [x'_m;x'_{m+1};...;x'_{m+l_{cf}-1}] \in \mathbb{R}^{d \cdot l_{cf}}$:
\begin{equation}
{h_{cf}} = f\left( {{W_{cf}} \cdot {x'_c} + b_{cf}} \right)
\label{eq:convolution}
\end{equation}
where $m$ is the index of words; $f$ is a non-linear activation function (we use $\tanh$\footnote{Some
papers suggest using $ReLU$ as the activation function in deep CNNs with
many layers. Nevertheless, we use $\tanh$ as the activation function, 
as our model is moderately deep and empirically we found the impact to be 
limited.}); $W_{cf} \in
\mathbb{R}^{len \times d \cdot {l_{cf}}}$ is the convolutional filter with
input length $d \cdot {l_{cf}}$ and output length $len$, where $l_{cf}$ is the
window size of the convolutional operation; and $h_{cf}$ and $b_{cf}$ are the
output and bias of the convolution layer $cf$, respectively. In our
experiments, the three window sizes $l_{cf}$ in the three convolution layers
are one, two, and three, encoding unigram, bigram, and trigram semantics
accordingly.

After the convolutional layer, we add a maximum pooling layer among convolutional
outputs to obtain the unigram, bigram, and trigram \textit{n}-gram representations.
This is succeeded by an average pooling layer for an element-wise average of
the three maximized convolution outputs. 

\subsection{UTCNN Model Description}
\label{sec:model}

\begin{figure}[tb]
\centering
\includegraphics[clip, trim=0.0cm 9.6cm 0.6cm 1.2cm, width=0.95\textwidth]{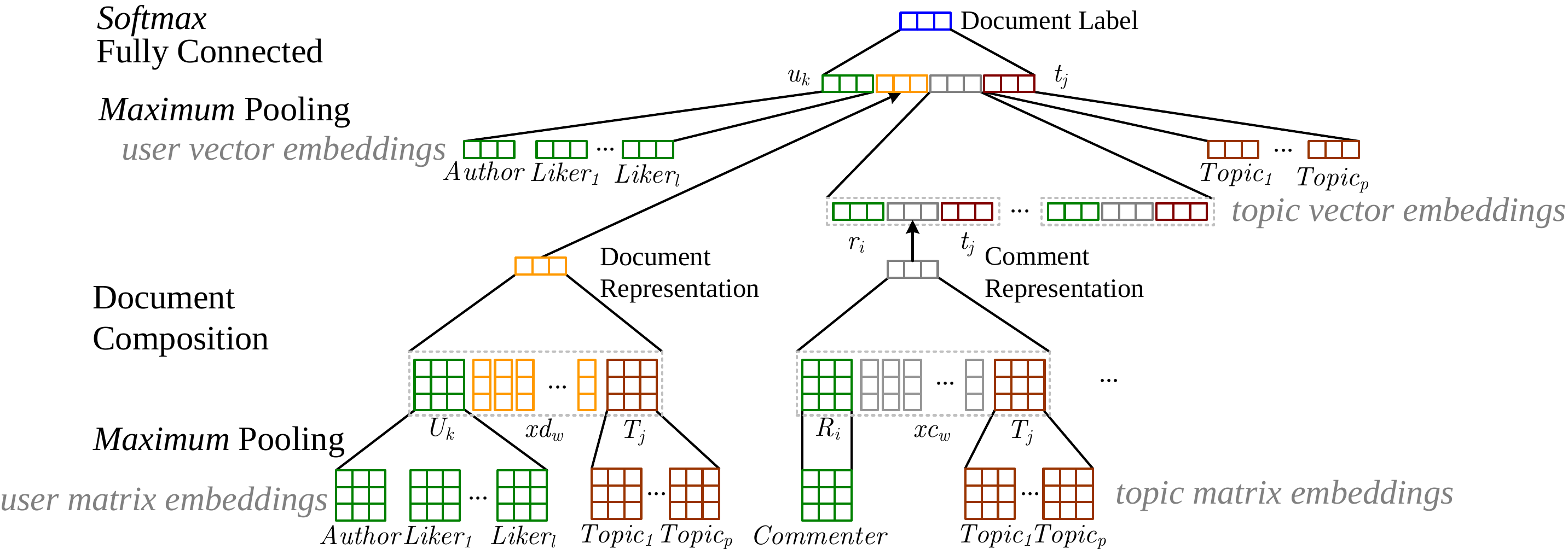}
\caption{The UTCNN model. Assuming one post author, $l$ likers and $p$ topics,
$xd_w$ is the word embedding of word $w$ in the document; $xc_w$ is the word
embedding of word $w$ in the comments; $U_k$ and $u_k$ are the moderator matrix
and vector embedding for moderator $k$; $T_j$ and $t_j$ are the topic matrix and
vector embedding for topic $j$; $R_i$ and $r_i$ are the commenter matrix and
vector embedding for commenter $i$. For simplicity we do not explicitly plot
the topic vector embedding part for comments, but it does include a maximum
pooling layer as with documents.}
\label{fig:utcnn}
\end{figure}

Figure~\ref{fig:utcnn} illustrates the UTCNN model. As more than one user
may interact with a given post, we first add a maximum pooling layer
after the user matrix embedding layer and user vector embedding layer to form a
moderator matrix embedding $U_k$ and a moderator vector embedding $u_k$ for
moderator $k$ respectively, where $U_k$ is used for the semantic transformation
in the document composition process, as mentioned in the previous section. 
The term \textit{moderator} here is to denote the pseudo user who provides the overall semantic/sentiment of all the engaged users for one document. 
The embedding $u_k$ models the moderator stance preference, that is, the pattern
of the revealed user stance: whether a user is willing to show his preference,
whether a user likes to show impartiality with neutral statements and
reasonable arguments, or just wants to show strong support for one stance.
Ideally, the latent user stance is modeled by $u_k$ for each user.
Likewise, for topic information, a maximum pooling layer is added after
the topic matrix embedding layer and topic vector embedding layer to form a
joint topic matrix embedding $T_j$ and a joint topic vector embedding $t_j$ for
topic $j$ respectively, where $T_j$ models the semantic transformation of
topic $j$ as in users and $t_j$ models the topic stance tendency. The latent
topic stance is also modeled by $t_j$ for each topic.

As for comments, we view them as short documents with authors only but without
likers nor their own comments\footnote{Recently Facebook released a function
allowing likes and comments on comments, but it was not available during the
time we collected data. However, UTCNN works on this richer data, as comments
are treated as posts under this framework.}. Therefore we apply document
composition on comments although here users are commenters (users who comment). 
It is noticed that the word embeddings $x_w$ for the same word in the posts and comments are the same, but after being transformed to $x'_w$ in the document composition process shown in Figure~\ref{fig:dc}, they might become different because of their different engaged users. 
The output comment representation together with the commenter vector embedding
$r_i$ and topic vector embedding $t_j$ are concatenated and a maximum
pooling layer is added to select the most important feature for comments.
Instead of requiring that the comment stance agree with the post,
UTCNN simply extracts the most important features of the comment contents; 
they could be helpful, whether they show obvious agreement or disagreement.
Therefore when combining comment information here, the maximum pooling layer 
is more appropriate than other pooling or merging layers. Indeed, we believe this
is one reason for UTCNN's performance gains.

Finally, the pooled comment representation, together with user vector embedding
$u_k$, topic vector embedding $t_j$, and document representation are fed to a
fully connected network, and softmax is applied to yield the final
stance label prediction for the post.

\section{Experiment}

\begin{table}[]
\centering
\begin{tabular}{l|| P{1.7em} P{2.7em} P{1.2em} P{2.7em} | P{1.6em} P{1.6em} P{1.6em} P{1.6em} P{1.6em} P{1.6em} P{1.6em} P{1.6em}}
\toprule
Dataset & 
\multicolumn{4}{c|}{FBFans} & 
\multicolumn{8}{c}{CreateDebate} 
\\ \midrule
\multirow{2}{*}{Type} & 
\multirow{2}{*}{\textit{Sup}} & 
\multirow{2}{*}{\textit{Neu}} &
\multirow{2}{*}{\textit{Uns}} & 
\multirow{2}{*}{All} & 
\multicolumn{2}{c}{ABO} &
\multicolumn{2}{c}{GAY} &
\multicolumn{2}{c}{OBA} &
\multicolumn{2}{c}{MAR}
\\ 
\cline{6-13}
 & & & & & 
{\textit{F}} & {\textit{A}} &
{\textit{F}} & {\textit{A}} &
{\textit{F}} & {\textit{A}} &
{\textit{F}} & {\textit{A}}
\\
\midrule
Training        & 7,097 & 19,412    & 245   & 26,754  
& 770.4 & 622.4 & 700.8 & 400.0 & 420.8 & 367.2 & 355.2 & 145.6
\\
Development     & 155   & 2,785     & 11    & 2,951     
& - & - & - & - & - & - & - & - 
\\
Testing         & 252   & 2,619     & 19    & 2,890       
& 192.6 & 155.6 & 175.2 & 100.0 & 105.2 & 91.8 & 88.8 & 36.4 
\\
\midrule
All             & 7,504 & 24,816    & 275   & 32,595      
& 963.0 & 778.0 & 876.0 & 500.0 & 526.0 & 459.0 & 444.0 & 182.0
\\
\bottomrule
\end{tabular}
\caption{Annotation results of FBFans and CreateDebate dataset.}
\label{tb:label}
\end{table}

\begin{table}[]
\centering
\label{tb:like}
\begin{tabular}{l|c | c | c }
\toprule
\diagbox{Author}{Post} & \textit{Sup} & \textit{Neu} & \textit{Uns} \\
\midrule
\textit{Sup} & \textbf{58.5\%} & \textbf{51.3}\% & 29.4\% \\
\textit{Neu} & 33.9\% & 43.5 \% & 9.3\% \\
\textit{Uns} & 7.6\% & 5.2\% & \textbf{61.3}\% \\
\bottomrule
\end{tabular}
\caption{Distribution of like behavior.}
\end{table}

We start with the experimental dataset and then describe the training process as well as
the implementation of the baselines. We also implement several variations to
reveal the effects of features: authors, likers, comment, and commenters. In
the results section we compare our model with related work.

\subsection{Dataset}

We tested the proposed UTCNN on two different datasets: FBFans and
CreateDebate. FBFans is a privately-owned\footnote{Currently not
released due to copyright and privacy issues.}, single-topic, Chinese,
unbalanced, social media dataset, and CreateDebate is a public,
multiple-topic, English, balanced, forum dataset. Results using these two
datasets show the applicability and superiority for different topics, languages,
data distributions, and platforms.  

The FBFans dataset contains data from anti-nuclear-power Chinese Facebook
fan groups from September 2013 to August 2014, including posts and their author
and liker IDs. There are a total of 2,496 authors, 505,137 likers, 33,686
commenters, and 505,412 unique users. 
Two annotators were asked to take into account only the post content to label
the stance of the posts in the whole dataset as \textit{supportive},
\textit{neutral}, or \textit{unsupportive} (hereafter denoted as \textit{Sup},
\textit{Neu}, and \textit{Uns}).
\textit{Sup}/\textit{Uns} posts were those in support of or against 
anti-reconstruction; \textit{Neu} posts were those evincing a neutral
standpoint on the topic, or were irrelevant. Raw agreement between annotators
is 0.91, indicating high agreement. Specifically, Cohen’s Kappa
for \textit{Neu} and not \textit{Neu} labeling is 0.58 (moderate), and
for \textit{Sup} or \textit{Uns} labeling is 0.84 (almost perfect). Posts with
inconsistent labels were filtered out, and the development and testing sets
were randomly selected from what was left. Posts in the development and testing sets involved
at least one user who appeared in the training set. 
The number of posts for each stance is shown on the left-hand side of
Table~\ref{tb:label}. About twenty percent of the posts were labeled with a
stance, and the number of supportive (\textit{Sup}) posts was much larger than
that of the unsupportive (\textit{Uns}) ones: this is thus highly skewed data, which
complicates stance classification. On average, 161.1 users were involved in one
post. The maximum was 23,297 and the minimum was one (the author). For
comments, on average there were 3 comments per post. The maximum was
1,092 and the minimum was zero.

To test whether the assumption of this paper – posts attract users who hold the same stance to like them – is reliable, we examine the likes from authors of different stances. Posts in FBFans dataset are used for this analysis. We calculate the like statistics of each distinct author from these 32,595 posts. As the numbers of authors in the \textit{Sup}, \textit{Neu} and \textit{Uns} stances are largely imbalanced, these numbers are normalized by the number of users of each stance. Table~\ref{tb:like} shows the results. Posts with stances (i.e., not neutral) attract users of the same stance. Neutral posts also attract both supportive and neutral users, like what we observe in supportive posts, but just the neutral posts can attract even more neutral likers. These results do suggest that users prefer posts of the same stance, or at least posts of no obvious stance which might cause annoyance when reading, and hence support the user modeling in our approach.

The CreateDebate dataset was collected from an English online debate
forum\footnote{\url{http://www.createdebate.com/}} discussing four topics: abortion
(ABO), gay rights (GAY), Obama (OBA), and marijuana (MAR). The posts are
annotated as \emph{for} (F) and \emph{against} (A). Replies to posts in this
dataset are also labeled with stance and hence use the same data format as
posts. The labeling results are shown in the right-hand side of
Table~\ref{tb:label}. We observe that the dataset is more balanced than the FBFans
dataset. In addition, there are 977 unique users in the dataset. To compare with
Hasan and Ng's work, we conducted five-fold cross-validation and present the
annotation results as the average number of all folds
\cite{hasan2013stance,hasan2014you}.

The FBFans dataset has more 
integrated functions than the CreateDebate dataset; 
thus our model can utilize all linguistic and extra-linguistic features.
For the CreateDebate dataset, on the other hand, the like and comment 
features are not available (as there is a stance label for each reply,
replies are evaluated as posts as other previous work)
but we still implemented our model using the content, author, and topic
information.

\subsection{Settings}
In the UTCNN training process, cross-entropy was used as the loss function and
\textit{AdaGrad} as the optimizer. For FBFans dataset, we learned the
50-dimensional word embeddings on the whole dataset using 
GloVe\footnote{\url{http://nlp.stanford.edu/projects/glove/}} \cite{glove} to
capture the word semantics; for CreateDebate dataset we used the
publicly available English 50-dimensional word embeddings, pre-trained also using
GloVe. These word embeddings
were fixed in the training process. The learning rate was set to 0.03. All user
and topic embeddings were randomly initialized in the range of [-0.1 0.1].
Matrix embeddings for users and topics were sized at 250 ($5 \times 50$);
vector embeddings for users and topics were set to length 10. 

We applied the LDA topic model \cite{blei2003latent} on the FBFans dataset to determine the latent
topics with which to build topic embeddings, as there is only one general known topic:
nuclear power plants. We learned 100 latent topics and assigned
the top three topics for each post. For the CreateDebate dataset, which itself 
constitutes four topics, the topic labels for posts were used directly without additionally
applying LDA.

For the FBFans data we report class-based f-scores as well as the
macro-average f-score ($\text{F}_\text{1}^\text{SNU}$) shown in equation~\ref{eq:macro}. 

\begin{eqnarray}
F_1^{SNU} = 2 \cdot \frac{{{P^{SNU}} \cdot {R^{SNU}}}}{{{P^{SNU}} + {R^{SNU}}}}
\label{eq:macro}
\end{eqnarray}
where $P^{SNU}$ and $R^{SNU}$ are the average precision and recall of the three class.
We adopted the macro-average f-score as the evaluation metric for the overall performance because (1)
the experimental dataset is severely imbalanced, which is
common for contentious issues; and (2) for stance classification, 
content in minor-class posts is usually more important for further
applications. For the CreateDebate dataset, accuracy was adopted as the
evaluation metric to compare the results with related work
\cite{hasan2013extra,hasan2013stance,sridhar2015joint}.

\subsection{Baselines}
We pit our model against the following baselines: 1) SVM with unigram, bigram,
and trigram features, which is a standard yet rather strong classifier for text
features; 2) SVM with average word embedding, where a document is represented as a
continuous representation by averaging the embeddings of the composite words; 3) SVM with average transformed word embeddings (the $x'w$ in equation \ref{eq:dot}), where a document is represented as a
continuous representation by averaging the transformed embeddings of the composite words; 4) two mature deep learning models on text classification, CNN \cite{CNN} and Recurrent Convolutional Neural Networks (RCNN) \cite{lai2015recurrent}, where the hyperparameters are based on their work; 
5) the above SVM and deep learning models with comment information; 
6) UTCNN without user information,
representing a pure-text CNN model where we use the same user matrix and user
embeddings $U_k$ and $u_k$ for each user; 
7) UTCNN without the LDA model,
representing how UTCNN works with a single-topic dataset; 
8) UTCNN without comments, in
which the model predicts the stance label given only user and topic
information. 
All these models were trained on the training set, and parameters as well as the SVM kernel selections (linear or RBF)
were fine-tuned on the development set. Also, we adopt oversampling on SVMs, CNN and RCNN because the FBFans dataset is highly imbalanced.


\subsection{Results on FBFans Dataset}

\begin{table}[t]
\centering
\begin{tabular}{l|| c c c c | P{1.6em} P{1.6em} P{1.6em} c}
\toprule
\multirow{2}{*}{\textbf{Method}} & 
 \multicolumn{4}{c|}{Features} & 
\multicolumn{3}{c}{F-score} & 
\multirow{2}{*}{$\text{F}_\text{1}^\text{SNU}$} \\ \cline{2-8} 
&Content & User & Topic &  Comment & \textit{Sup}    & \textit{Neu}    & \textit{Uns}   &                       \\ 
\midrule
Majority & & & & & .000   & .841   & .000  & \mbox{.280}
\\

\mbox{SVM
-UniBiTrigram}& ${\surd}$  & & & & .721   & .967   & .091  & \mbox{.640}   \\ 
\mbox{SVM
-UniBiTrigram}& ${\surd}$ & &  & ${\surd}$ & .610   & .938   & .156  & \mbox{.621}   \\ 
\mbox{SVM
-AvgWordVec}& ${\surd}$  & & & & .631   & .952   & .114  & \mbox{.579}   \\ 
\mbox{SVM
-AvgWordVec}& ${\surd}$ &   &  & ${\surd}$ & .526   & .100   & .165  & \mbox{.336}   
\\
\mbox{SVM
-AvgWordVec (transformed)}& ${\surd}$ & ${\surd}$ & ${\surd}$ & & .571  & .920   & .229  & \mbox{.637}   \\ 
\mbox{SVM
-AvgWordVec (transformed)}& ${\surd}$ & ${\surd}$ & ${\surd}$ & ${\surd}$ & .597   & .963   & .210  & \mbox{.642}   
\\
\midrule
\mbox{CNN \cite{CNN}}& ${\surd}$  & & & & .738   & .967   & .171  & \mbox{.637}   \\ 
\mbox{CNN \cite{CNN}}& ${\surd}$ &   &  & ${\surd}$ & .726   & .964   & .222  & \mbox{.648}   
\\
\mbox{RCNN \cite{lai2015recurrent}}& ${\surd}$ & & & & .669   & .951   & .079  & \mbox{.606}   \\ 
\mbox{RCNN \cite{lai2015recurrent}}& ${\surd}$ & & & ${\surd}$ & .628   & .944   & .096  & \mbox{.605}   
\\
\midrule
UTCNN without user& ${\surd}$ &   & ${\surd}$ &  ${\surd}$   & \textbf{.748}   & \textbf{.973}   & .000  & \mbox{.580} \\
UTCNN without topic& ${\surd}$  & ${\surd}$ &       & ${\surd}$        & .643   & .944   & .476  & \mbox{.706} \\
UTCNN without comment& ${\surd}$  & ${\surd}$ & ${\surd}$ &     & .632   & .940   & .480  & \mbox{.707} \\
UTCNN shared user embedding & ${\surd}$  & ${\surd}$ & ${\surd}$ & ${\surd}$  & .625   & .969   & .531  & \mbox{.732} 
\\
UTCNN (full) & ${\surd}$  & ${\surd}$ & ${\surd}$ & ${\surd}$ & .698   & .957   & \textbf{.571}  & \mbox{\textbf{.755}*} \\
\bottomrule
\end{tabular}
\caption{Performance of post stance classification on the FBFans dataset. \\
{\footnotesize *UTCNN (full) results are statistically 
significant ($p$-value $<$ 0.005) with respect to all other methods 
except for UTCNN shared user embedding.}}
\label{tb:performance1}
\end{table}

In Table~\ref{tb:performance1} we show the results of UTCNN and the baselines
on the FBFans dataset. Here Majority yields good performance on \textit{Neu} since
FBFans is highly biased to the neutral class. The SVM models perform well on
\textit{Sup} and \textit{Neu} but perform poorly for \textit{Uns}, showing that
content information in itself is insufficient to predict stance labels,
especially for the minor class. 
With the transformed word embedding feature, SVM can achieve comparable performance as SVM with \textit{n}-gram feature. However, the much fewer feature dimension of the transformed word embedding makes SVM with word embeddings a more efficient choice for modeling the large scale social media dataset. For the CNN and RCNN models, they perform slightly better than most of the SVM models but still, the content information is insufficient to achieve a good performance on the \textit{Uns} posts.
As to adding comment information to these models, since the commenters do not always
hold the same stance as the author, simply adding comments and post contents
together merely adds noise to the model.

Among all UTCNN variations, we find that user information is most important,
followed by topic and comment information. UTCNN without user information
shows results similar to SVMs --- it does well for \textit{Sup} and
\textit{Neu} but detects no \textit{Uns}. Its best f-scores on both
\textit{Sup} and \textit{Neu} among all methods show that with enough
training data, content-based models can perform well; at the same time, the
lack of user information results in too few clues for minor-class posts
to either predict their stance directly or link them to other users and
posts for improved performance. 
The 17.5\%
improvement when adding user information suggests that user information is
especially useful when the dataset is highly imbalanced. All models that
consider user information predict the minority class successfully. 
UCTNN without topic information works well but achieves lower performance than the full UTCNN model. The 4.9\% 
performance gain brought by LDA shows that
although it is satisfactory for
single topic datasets, adding that latent topics still benefits performance: even when we are discussing the same topic, we use
different arguments and supporting evidence.
Lastly, we get 4.8\% improvement
when adding comment information and it achieves comparable performance to UTCNN without topic
information, which shows that comments also benefit performance. For platforms where user IDs are pixelated or otherwise hidden, adding comments to a text
model still improves performance.
In its integration of user, content, and comment information, the full
UTCNN produces the highest f-scores on all \textit{Sup}, \textit{Neu}, and
\textit{Uns} stances among models that predict the \textit{Uns} class, and the
highest macro-average f-score overall. This shows its ability to balance a
biased dataset and supports our claim that UTCNN successfully bridges content and
user, topic, and comment information for stance classification on social media
text. Another merit of UTCNN is that it does not require a balanced training data. This is supported by its outperforming other models though no oversampling technique is applied to the UTCNN related experiments as shown in this paper. Thus we can conclude that the user information provides strong clues and it is still rich even in the minority class.

We also investigate the semantic difference when a user acts as an
author/liker or a commenter. We evaluated a variation in which all embeddings from
the same user were forced to be identical (this is the UTCNN shared user embedding 
setting in Table~\ref{tb:performance1}). 
This setting yielded only a 2.5\% improvement over the model without
comments, which is not statistically significant. 
However, when separating authors/likers and commenters embeddings (i.e., the
UTCNN full model), we achieved much greater improvements (4.8\%). We
attribute this result to the tendency of users to use different wording for
different roles (for instance author vs commenter). This is observed
when the user, acting as an author, attempts to support her argument against
nuclear power by using improvements in solar power;
when acting as a commenter, though, she interacts with post contents by
criticizing past politicians who supported nuclear power or by
arguing that the proposed evacuation plan in case of a nuclear accident is ridiculous.
Based on this finding, in the final UTCNN setting we train two user matrix
embeddings for one user: one for the author/liker role and the other for
the commenter role.

\subsection{Results on CreateDebate Dataset}

\begin{table}[t]
\centering
\begin{tabular}{l|| c c | c c c c c}
\toprule
\multirow{2}{*}{\textbf{Method}} & \multicolumn{2}{c|}{Features} &  \multicolumn{4}{c}{Topics} & \multirow{2}{*}{AVG}
\\
\cline{2-7}
& Text & User & ABO & GAY & OBA & MAR  &
\\ 
\midrule
Majority    &       &         & .549   & .634   & .539  & .695 & .604 \\
\mbox{SVM
-UniBiTrigram} & ${\surd}$ &   & .592   & .569   & .565  & .673 & .600
\\
\mbox{SVM
-AvgWordVec} & ${\surd}$ &   & .559   & .637   & .548  & .708 & .613
\\
\mbox{SVM
-AvgWordVec (transformed)} & ${\surd}$ & ${\surd}$ & .859   & .830   & .800  & .741 & .808
\\
\midrule
\mbox{CNN \cite{CNN}} & ${\surd}$ &   & .553   & .636   & .557  & .709 & .614
\\
\mbox{RCNN \cite{lai2015recurrent}} & ${\surd}$ &   & .553   & .637   & .534  & .709 & .608
\\
\midrule
ILP \cite{hasan2013extra} & ${\surd}$ &   & .614   & .626   & .581  & .669 & .623
\\
ILP \cite{hasan2013extra}& ${\surd}$ & ${\surd}$ & .749   & .709   & .727  & .754 & .735
\\
CRF \cite{hasan2013stance} & ${\surd}$ & ${\surd}$   & .747   & .699   & .711  & .754 & .728
\\
PSL \cite{sridhar2015joint} & ${\surd}$ & ${\surd}$   & .668   & .727   & .635  & .690 & .680
\\
\midrule
UTCNN without topic   & ${\surd}$ & ${\surd}$      & .824   & \textbf{.851}   & .743  & \textbf{.814} & .808
\\
UTCNN without user  & ${\surd}$ &       & .617   & .627  
& .599  & .685 & .632
\\
UTCNN (full)     & ${\surd}$ & ${\surd}$         & \textbf{.878}   & .850   & \textbf{.857}  & .782  & \textbf{.842*}
\\
\bottomrule
\end{tabular}
\caption{Accuracies of post stance classification on CreateDebate dataset. \\
{\footnotesize *UTCNN results were statistically significant ($p$-value $<$ 0.001) with 
respect to other UTCNN settings.}}
\label{tb:performance2}
\end{table}

Table~\ref{tb:performance2} shows the results of UTCNN, baselines as we implemented on the FBFans datset and related work on
the CreateDebate dataset. We do not adopt oversampling on these models because the CreateDebate dataset is almost balanced. In previous work, integer linear programming (ILP) or
linear-chain conditional random fields (CRFs) were proposed to integrate text
features, author, ideology, and user-interaction constraints, where text
features are unigram, bigram, and POS-dependencies; the author constraint tends to
require that posts from the same author for the same topic hold the same stance;
the ideology constraint aims to capture inferences between topics for the same
author; the user-interaction constraint models relationships among posts via
user interactions such as replies \cite{hasan2013extra,hasan2013stance}. 

The SVM with \textit{n}-gram or average word embedding feature performs just similar to the majority. However, with the transformed word embedding, it achieves superior results. It shows that the learned user and topic embeddings really capture the user and topic semantics. This finding is not so obvious in the FBFans dataset and it might be due to the unfavorable data skewness for SVM. As for CNN and RCNN, they perform slightly better than most SVMs as we found in Table~\ref{tb:performance1} for FBFans.

Compared to the ILP \cite{hasan2013extra} and CRF \cite{hasan2013stance}
methods, the UTCNN user embeddings encode author and user-interaction
constraints, where the ideology constraint is modeled by the topic embeddings
and text features are modeled by the CNN. The significant improvement achieved by UTCNN
suggests the latent representations are more effective than overt model
constraints.

The PSL model \cite{sridhar2015joint} jointly labels both author
and post stance using probabilistic soft logic (PSL) \cite{bach2015hinge} by
considering text features and reply links between authors and posts as in Hasan
and Ng's work. Table~\ref{tb:performance2} reports the result of their best 
AD setting, which represents the full joint stance/disagreement
collective model on posts and is hence more relevant to UTCNN. 
In contrast to their model, the UTCNN user embeddings represent
relationships between authors, but UTCNN models do not utilize link information 
between posts. Though the PSL model has the advantage of being able to jointly label
the stances of authors and posts, its performance on posts is lower
than the that for the ILP or CRF models. UTCNN significantly
outperforms these models on posts and has the potential to predict user stances
through the generated user embeddings. 

For the CreateDebate dataset, we also evaluated performance when not 
using topic embeddings or user embeddings; 
as replies in this dataset are viewed as posts, 
the setting without comment embeddings is not available. 
Table~\ref{tb:performance2} shows the same findings as
Table~\ref{tb:performance1}: the 21\% improvement in accuracy demonstrates that user 
information is the most vital. This finding also supports the results in the
related work: user constraints are useful and can yield 11.2\% improvement in
accuracy \cite{hasan2013extra}. Further considering topic information yields
3.4\% improvement, suggesting that knowing the subject of debates provides
useful information. In sum, Table~\ref{tb:performance1} together with
Table~\ref{tb:performance2} show that UTCNN achieves promising performance
regardless of topic, language, data distribution, and platform.

\section{Conclusion}
We have proposed UTCNN, a neural network model that incorporates user, topic,
content and comment information for stance classification on social media
texts. UTCNN learns user embeddings for all users with minimum active
degree, i.e., one post or one like. 
Topic information obtained from the topic model or the pre-defined labels
further improves the UTCNN model. In addition, comment information
provides additional clues for stance classification.
We have shown that UTCNN achieves promising and balanced results. In the future
we plan to explore the effectiveness of the UTCNN user embeddings for
author stance classification.

\section*{Acknowledgements}
Research of this paper was partially supported by Ministry
of Science and Technology, Taiwan, under the contract
MOST 104-2221-E-001-024-MY2.

\bibliographystyle{coling2016}
\bibliography{coling2016}

\end{document}